\pdfoutput=1
\documentclass[letterpaper, 10 pt, conference]{ieeeconf}  

\IEEEoverridecommandlockouts                              

\usepackage{amsmath}
\usepackage{amssymb}
\usepackage{graphicx}
\usepackage{xcolor}
\usepackage{hyperref}

\usepackage[top=57pt,bottom=43pt,left=48pt,right=48pt]{geometry}

\author{Hrishikesh Gupta$^{1}$, Stefan Thalhammer$^{2}$, Jean-Baptiste Weibel$^{1}$, Alexander Haberl$^{1}$ and Markus Vincze$^{1}$
\thanks{*The final version of this paper is published in IEEE Robotics and Automation Letters (DOI: 10.1109/LRA.2024.3455897). This work was supported by the EU-program EC Horizon 2020 for Research and Innovation under grant agreement No. 101017089, project TraceBot.}
\thanks{$^{1}$Hrishikesh Gupta, Jean-Baptiste Weibel, Markus Vincze, and Alexander Haberl are with Vision for Robotics Laboratory, 
Automation and Control Institute, 
TU Wien, Austria {\tt\small \{gupta, weibel, vincze, haberl\}@acin.tuwien.ac.at}}%
\thanks{$^{2}$Stefan Thalhammer is with the Industrial Engineering Department, UAS Technikum Vienna, 
TU Wien, Austria {\tt\small stefan.thalhammer@technikum-wien.at}}%
}

\begin{document}
\title{\LARGE \bf
ReFlow6D: Refraction-Guided Transparent Object 6D Pose Estimation via Intermediate Representation Learning
}

\maketitle
\thispagestyle{empty}
\pagestyle{empty}

\begin{abstract}
Transparent objects are ubiquitous in daily life, making their perception and robotics manipulation important. However, they present a major challenge due to their distinct refractive and reflective properties when it comes to accurately estimating the 6D pose. To solve this, we present \textit{ReFlow6D}, a novel method for transparent object 6D pose estimation that harnesses the \textit{refractive-intermediate representation}. Unlike conventional approaches, our method leverages a feature space impervious to changes in RGB image space and independent of depth information. Drawing inspiration from image matting, we model the deformation of the light path through transparent objects, yielding a unique object-specific intermediate representation guided by light refraction that is independent of the environment in which objects are observed. By integrating these intermediate features into the pose estimation network, we show that \textit{ReFlow6D} achieves precise 6D pose estimation of transparent objects, using only RGB images as input. Our method further introduces a novel transparent object compositing loss, fostering the generation of superior \textit{refractive-intermediate} features. Empirical evaluations show that our approach significantly outperforms state-of-the-art methods on \textit{TOD} and \textit{Trans32K-6D} datasets. Robot grasping experiments further demonstrate that \textit{ReFlow6D's} pose estimation accuracy effectively translates to real-world robotics task.
The source code is available at: \href{https://github.com/StoicGilgamesh/ReFlow6D}{https://github.com/StoicGilgamesh/ReFlow6D} and \href{https://github.com/StoicGilgamesh/matting\_rendering}{https://github.com/StoicGilgamesh/matting\_rendering}.
\end{abstract}

\section{Introduction} \label{seq:intro}

Robot object manipulation using 6D pose estimation is a well-studied problem in robotics, e.g., \cite{thalhammer2021pyrapose}\cite{stevvsic2020learning}. It uses the estimation of the 6D pose of objects, i.e., 3D rotation and 3D translation, for solving a wide variety of real-world tasks such
as object grasping, scene understanding, and complex robotic object manipulation.
Researchers have extensively studied this problem for opaque objects \cite{hodavn2016evaluation} such as texture-less objects \cite{hodan2017t},\cite{wu2021pseudo}, symmetrical objects \cite{hodan2020epos}\cite{richter2021handling} and also occluded objects \cite{peng2019pvnet}. However, even though transparent objects are as pervasive in daily life and household settings as opaque objects, their 6D pose estimation has been approached only in comparably few works\cite{chang2021ghostpose}\cite{lysenkov2013pose}.

Methods developed for 6D pose estimation of opaque and Lambertian objects cannot be directly applied to transparent objects without degradation of performance. This is mainly because transparent objects pose two prominent challenges to visual perception system. Firstly, transparent objects do not exhibit consistent RGB color and texture features across varying
scenes. A transparent object's appearance depends on the scene’s lighting, background, and setup. Hence, the visual features drastically differ between scenes thereby confounding RGB feature-based learning methods. 
Secondly, the non-Lambertian nature of transparent objects poses challenges for commercial depth sensors and leads to inaccurate depth measurements \cite{sajjan2020clear, chen2022clearpose}. This limitation significantly impacts state-of-the-art pose estimation approaches reliant on precise depth data. 

\begin{figure}\label{Fig1}
\centering
 \hfill\includegraphics[width=0.48\textwidth]{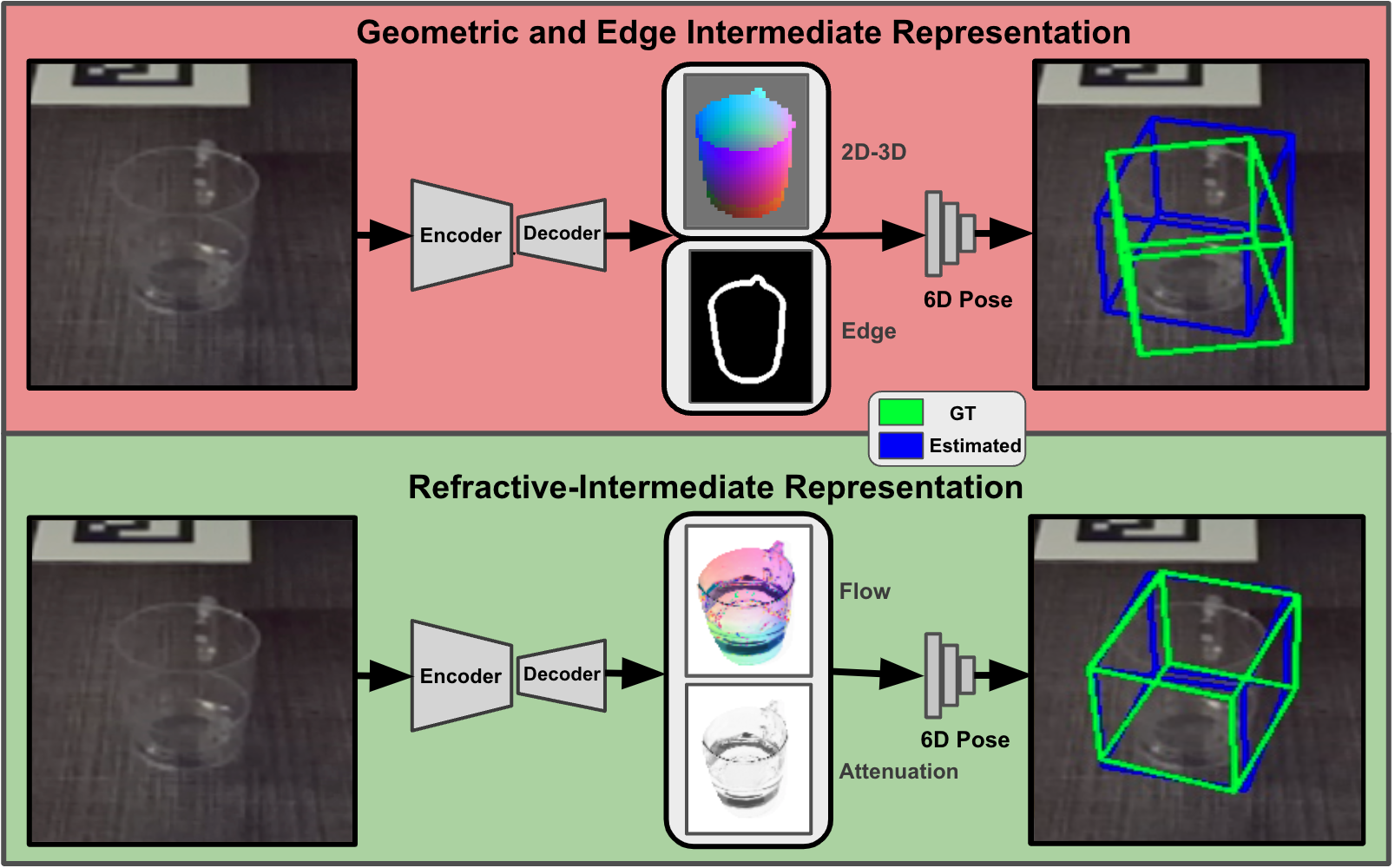}
 \caption{\textbf{Intermediate representation for pose estimation}: The figure shows the effectiveness of the refractive-intermediate representation vs Geometric and edge intermediate layers applied to 6D pose estimation. The green 3D Bbox shows groundtruth, while the blue shows the estimation.}  
\end{figure}

Research on transparent objects can be divided into the following primary directions. One approach focuses on completing missing or erroneous depth information for transparent objects \cite{zhu2021rgb, sajjan2020clear} followed by utilizing a secondary network for pose estimation \cite{chen2022clearpose}. The second approach focuses on estimating poses from implicit depth cues, such as stereo images \cite{liu2020keypose}. Both approaches leverage depth cues, yet recent advancements such as \cite{wang2021gdr} have demonstrated the effectiveness of estimating poses using the RGB image space only, while incorporating the relevant geometric features for opaque objects \cite{hodan2018bop}. Furthermore, \cite{yu2023tgf} has extended this methodology to transparent objects by showing the effectiveness of intermediate geometric and edge representations.

In this paper, we present \textit{ReFlow6D}, a novel approach for transparent object 6D pose estimation that leverages \textit{refractive-intermediate representation} (see Fig.~1), a more reliable intermediate feature space for transparent objects. This representation captures the deformation of the light path induced by the given transparent object. This deformation is consistent and independent of the environment and accounts for symmetries of the object. As a result, making \textit{ReFlow6D} robust to changes in RGB image space and operate independently of the depth information.

In detail, for a given pose, the deformation of light induced by a transparent object can be expressed as a refractive flow and an attenuation. The refractive flow is the offset between a foreground pixel and its refracted counterpart on the background, while the attenuation captures light intensity change at each pixel. These properties were utilized by Tom-Net \cite{chen2018tom} for transparent object matting and compositing. Image matting is the process of extracting a foreground object from a background image, yielding an opacity value (and refraction in the case of a transparent object) for each pixel of the extracted object referred to as \textit{matte}. Inversely, image compositing combines the foreground object with a new background,  guided by the extracted \textit{matte} to reveal or conceal specific regions \cite{zongker2023environment}. The refractive flow and attenuation as detailed before are unique optical properties of a transparent object regardless of the environment where it is placed. We combine these optical properties mentioned in \cite{chen2018tom} with the object binary mask and surface region attention maps (Surface Regions), forming our \textit{refractive-intermediate representation} for transparent objects. Using a learned Patch-PnP \cite{wang2021gdr} we directly regress the 6D object pose from these learned  \textit{refractive-intermediate representation}.

To summarize, the contributions towards solving transparent object 6D pose estimation for robot object handling are:
\begin{enumerate}
    \item Incorporation of the \textit{refractive-intermediate representation} in the pose estimation architecture as intermediate features that model the deformation of light paths through transparent objects, which is a unique matte for a transparent object and invariant to environment changes, enabling more robust and accurate 6D pose estimation.
    \item Evaluations against the state of the art showing the improvement of estimation of 6D poses of transparent objects using the refractive over the geometry intermediate representation.
    \item Robot transparent object manipulation experiments demonstrating the real applicability of \textit{ReFlow6D}.
\end{enumerate}

The paper is organized as follows: Section 2 covers related work. Section 3 explains ReFlow6D and the network architecture. Section 4 presents the experimental evaluation and results against state-of-art methods. Section 5 showcases robot grasping experiments in a real-world environment using the pose estimates of ReFlow6D. Finally, we conclude this paper in section 6.

\section{Related Works}

In this section, we briefly review representative and recent works
on the perception of the transparent objects, their manipulation and transparent object pose estimation and image matting. 

\begin{figure*}[htp]
    \centering
  \includegraphics[width=0.87\textwidth]{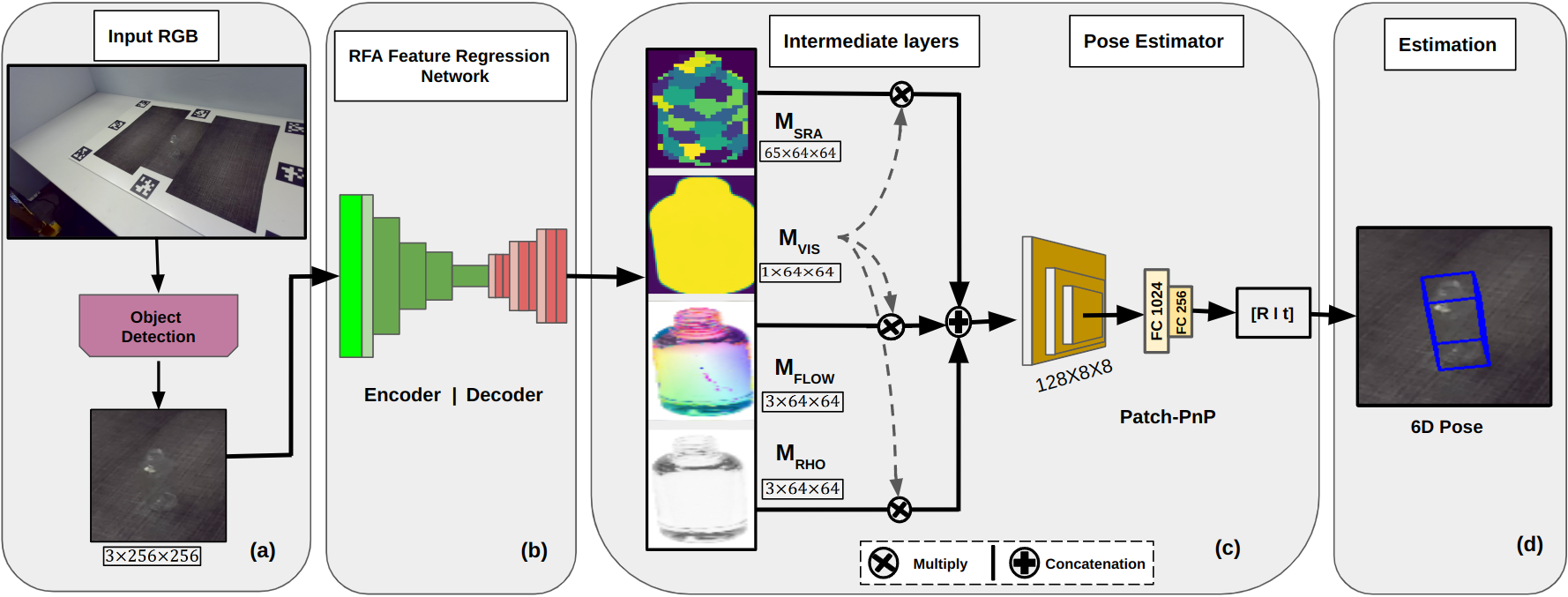}
  \caption{\textbf{Framework of ReFlow6D}: (a) Given an RGB image $I$ we use off-the-shelf object detector for detecting transparent objects. (b) The RFA feature regression network takes then zoomed-in RoI as input and predicts several refractive-intermediate representation. (c) These intermediate features are then concatenated and provided as input to the Patch-PnP. (d) The Patch-PnP directly regresses the 6D object pose of the transparent object.}
\end{figure*}

\subsection{Transparent object perception and manipulation}

Perception of transparent objects is a necessary precursor to robotic manipulation. In order to address the challenges associated with transparent objects, CNN models for the detection of transparent objects were introduced by \cite{lai2015transparent} using the RGB image space only. Further, \cite{xie2020segmenting} proposed a deep segmentation model that achieved state-of-the-art accuracy in segmentation tasks. Specialized sensors such as in \cite{kalra2020deep} training Mask R-CNN using polarization images to outperform baseline models have also been explored.

A solution to accurately measure the depth information of transparent objects is to estimate the missing depth through depth completion. In the context of pose estimation and robotic grasping, ClearGrasp \cite{sajjan2020clear} utilized depth completion techniques, training DeepLabv3+ models for image segmentation, surface normal estimation, and boundary segmentation before robot grasping. Further advancements in in-depth completion include methods involving implicit functions \cite{zhu2021rgb} and NeRF features \cite{ichnowski2021dex}.
\subsection{Transparent object 6D pose estimation and image matting}

Solving pose estimation problems using implicit depth cues, Keypose \cite{liu2020keypose} was introduced for regressing 2D keypoints using stereo images. It outperformed DenseFusion \cite{wang2019densefusion}, even with groundtruth depth. \cite{chang2021ghostpose} proposed a transparent object grasping approach estimating
the object 6D poses from the proposed model-free pose
estimation approach using multiview-geometry.

In the realm of monocular RGB-based methods, Transnet \cite{zhang2022transnet} directly regresses transparent object
pose from images using depth completion and surface normal estimation. GDR-Net \cite{wang2021gdr} unifies direct and geometry-based indirect methods, providing direct pose information output for opaque objects. Building upon CDPN \cite{li2019cdpn} and EPOS \cite{hodan2020epos} GDR-Net uses geometric features such as surface region attention and dense correspondence maps as intermediate layers, directly regressing 6D poses using their proposed Patch-PnP method. Further, building upon GDR-Net, TGF-Net \cite{yu2023tgf} regresses direct 6D poses with the Patch-PnP method using their proposed edge intermediate representations, demonstrating improvement over the proposed geometric representations presented in \cite{wang2021gdr} for transparent objects.

Object matting and compositing were detailed early by Zongker et al.\cite{zongker2023environment} and \cite{chuang2000environment}. Building on this TOM-Net \cite{chen2018tom} frame the transparent object image matting problem as a refractive flow and attenuation estimation problem. They show that this matte can be effectively estimated and learned by CNN-based methods for colorless transparent objects. Further, by compositing the transparent object on varying backgrounds using the learned matte, they also show that the extracted matte (RFA) is a unique property of the transparent object, independent of its environment.  

We aim to leverage this distinct matte to tackle transparent object pose estimation, over the geometric \cite{wang2021gdr} and edge intermediate representation \cite{yu2023tgf}.

\section{ReFlow6D}

We now present our method to solve the transparent object 6D pose estimation problem, given the input RGB image $I$ and a set of $N$ transparent objects $O$ =\{ $O_i$ $|$ i = 1..N \} together with their corresponding 3D CAD models $S$ = \{ $S_i$ $|$ i = 1..N \}. First, all transparent objects of interest from $O$ in the image $I$ are detected using a standard object detector like \cite{tian2019fcos, redmon2018yolov3}. 
We then aim to learn the \textit{refractive-intermediate representation} for the object $O_i$ in the image $I$. 

Using these learned intermediate representations we then perform the 6D pose regression $P_i$ = [$R_i|t_i$] w.r.t. the camera for each object $O_i$ present in $I$, where $R_i$ describes the 3D rotation and $t_i$ denotes the 3D translation of the detected object. Fig. 2 provides a schematic overview of the proposed ReFlow6D. We utilize the Patch-PnP method proposed by GDR-Net \cite{wang2021gdr}, enabling us to perform a direct pose regression from the concatenation of these learned intermediate features. This stands in contrast to earlier methods that employed indirect pose estimation approaches \cite{li2019cdpn}.

Following, Section~\ref{sec:rfa_comp} illustrates the proposed \textit{refractive-intermediate representation} and how it is estimated. Further in Section~\ref{sec:losses}, we detail the losses we employ to learn this intermediate representation and the 6D pose regression from the Patch-PnP method. Additionally, Section~\ref{sec:comp_loss} will provide details on the object compositing loss that we use for further supervision to refine the estimated \textit{refractive-intermediate representation}.

\subsection{Refractive-intermediate representation}\label{sec:rfa_comp}
In this subsection, we introduce in detail the intermediate representation we use for estimating a direct 6D pose regression using Patch-PnP. For learning the \textit{refractive-intermediate representation}, we employ an encoder-decoder architecture, specifically utilizing the Geometric feature regression network proposed in \cite{wang2021gdr}. This choice enables us to learn the intermediate representations with minimal adjustments to their network and directly regress poses using their proposed Patch-PnP. Specifically, we retain the layers for regressing the surface-region map and object mask while adding channels necessary for refractive flow and attenuation regression. We refer to this modified network as \textit{RFA regression network} (as seen in Fig. 2).

\textbf{Refractive flow and attenuation (RFA).} The refractive flow $M_{\text{FLOW}}$ represents the deformation of the light ray traversing through the transparent object from its background environment to the foreground object's surface. Where, each pixel on the refractive flow image is a 2D vector ($\delta x, \delta y$), which indicates the offset between the observed foreground pixel and its corresponding background pixel after the refraction of the ray~\cite{chen2018tom}. The attenuation or the attenuation index $M_{\text{RHO}}$, as mentioned in TOM-Net, represents the intensity of the light-ray traversing through the object on its surface. The attenuation represents the magnitude or light intensity at a pixel. The RFA of an object $ O_i \in O$ can be described as a function of index of refraction($IOR$) of the $ O_i$ and $S_i$. The parameter $S_i$ plays a crucial role in deriving the shape and geometry of the object. Since $IOR$, the shape, and geometry of the object $O_i$ are constant attributes, their RFA properties (refractive flow and attenuation) remain constant as well. These RFA properties of an object $ O_i \in O$ remain unchanged, regardless of the environment it is placed in. Forming a set of physically plausible intermediate features independent to changes in the environment of the transparent object. 

\textbf{Handling Object Symmetries with Surface Region Attention maps.} We implement Surface Regions $M_{\text{SRA}}$, akin to prior works \cite{wang2021gdr, yu2023tgf,su2022zebrapose}. These fragments encapsulate rich geometry information, while providing additional ambiguity-aware supervision for the 6D pose estimation. We predict $M_{\text{SRA}}$ similarly to \cite{wang2021gdr}, treating it as an intermediate layer. Inspired by \cite{hodan2020epos}, $M_{\text{SRA}}$ serves as a symmetry-aware feature map guiding Patch-PnP feature learning. $M_{\text{SRA}}$ derived from Dense Correspondence Maps by employing farthest points sampling. \cite{wang2021gdr}, classifying each pixel in the dense correspondence maps into the corresponding regions. Thus the classification probabilities predicted for each pixel in $M_{SRA}$ implicitly represent the symmetry of the object. 

In addition to the above mentioned intermediate representations, we also predict the mask of the object. This implicitly encodes the geometric information of the object in addition to $M_{\text{SRA}}$. Specifically, we predict the object visibility mask $M_{VIS}$ for each detected object in image $I$.

\begin{figure}[htbp]
    \centering
    \includegraphics[width=0.40\textwidth]{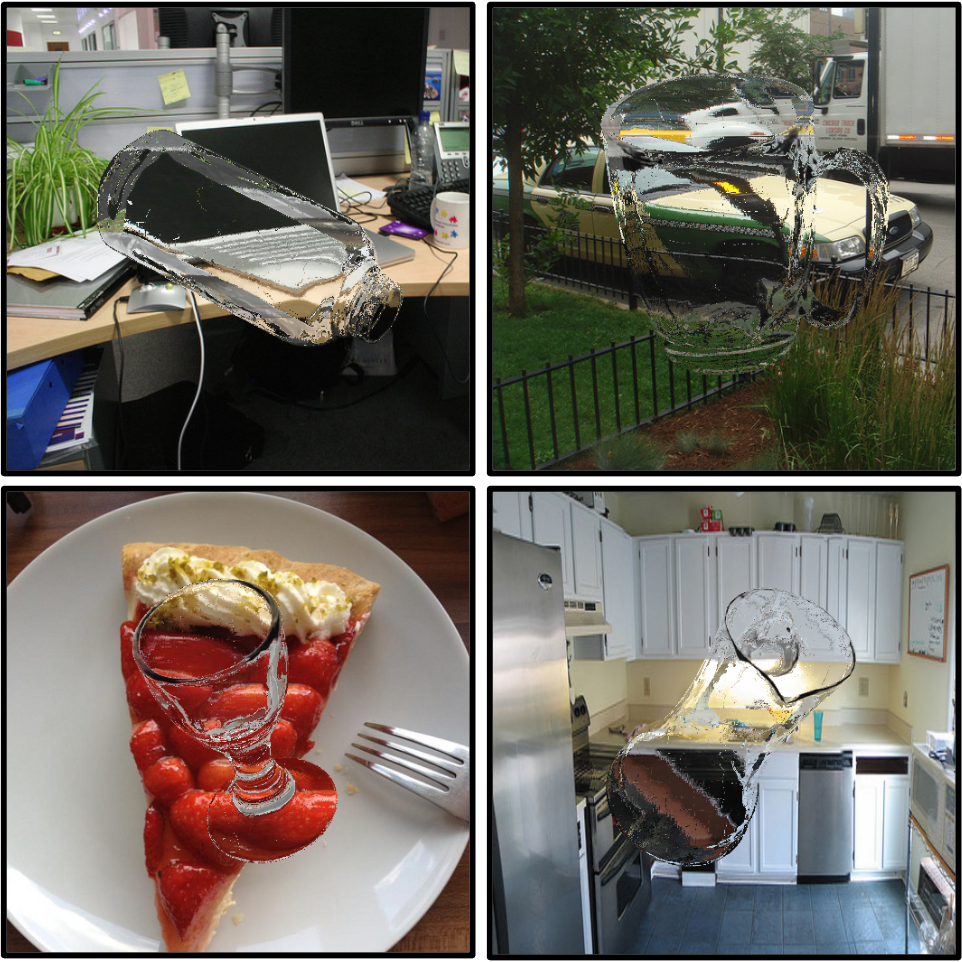}
    \caption{\textbf{Transparent object Compositing:} Examples of transparent object compositing from the TOD and Trans32K-6D datasets on random COCO backgrounds. Used for additional supervision loss for refining the estimated RFA.}
\end{figure}

\subsection{Loss Functions} \label{sec:losses}

The final loss $L$ function of our optimization scheme comprises two distinct loss functions:

\begin{equation}
 L = L_{\text{inter}} + L_{\text{pose}} \label{eq:FinalLoss}
\end{equation}

Where $L_{\text{inter}}$ denotes the loss function for our intermediate features and $L_{\text{pose}}$ represents loss for our output patch-PnP \cite{wang2021gdr} pose. $L_{\text{inter}}$ function can individually be denoted as:

\begin{equation}
 L_{\text{inter}} = L_{\text{FLOW}}+ L_{\text{RHO}} + L_{\text{M}_{\text{VIS}}}
\end{equation}
Where,

\begin{equation}
\left\{
\begin{aligned}
    & L_{\text{FLOW}} = || \hat{M}_{\text{VIS}} \cdot ( \tilde{M}_{\text{FLOW}} - \hat{M}_{\text{FLOW}} ) ||_1 \\
    & L_{\text{RHO}} = || \hat{M}_{\text{VIS}} \cdot ( \tilde{M}_{\text{RHO}} - \hat{M}_{\text{RHO}} ) ||_1 \\
    &  L_{\text{M}_{\text{VIS}}} = ||\tilde{M}_{\text{VIS}} - \hat{M}_{\text{VIS}} ||_1 \\   
    \end{aligned}
\right.
\end{equation}

Where, $\hat{\textbf{.}}$ and $\tilde{\textbf{.}}$ denote groundtruth and estimation, respectively. $L_{\text{pose}}$  function can individually be denoted as:
\begin{equation}
 L_{\text{pose}} = L_{\text{R}} + L_{\text{center}} + L_{\text{Z}} \label{eq:L_Pose}
\end{equation}

Where following \cite{wang2021gdr} we employ a disentangled 6D pose loss via individually supervising the rotation $R$, the scale-invariant 2D object center $(\delta_{x}, \delta_{y})$, and
the distance $\delta_{z}$. Where the components of eq. \ref{eq:L_Pose} are:

\begin{equation}
\left\{
\begin{aligned}
    & L_{\text{R}} = avg_{x \in S} || \hat{Rx} - \tilde{Rx} ||_1 \\
    & L_{\text{center}} = || \hat{\delta_x} - \tilde{\delta_x}, \hat{\delta_y} - \tilde{\delta_y} ||_1 \\
    & L_{\text{Z}} = || \hat{\delta_z} - \tilde{\delta_z} ||_1 \\ 
    \end{aligned}
\right.
\end{equation}

\textbf{Object compositing as additional supervision.} \label{sec:comp_loss}
The quality of predicted Refractive Flow and Attenuation (RFA) features as intermediate layers strongly influence the accuracy of the final 6D pose estimation for transparent objects, as demonstrated in subsequent ablation studies. Observations in \cite{chen2018tom} show that higher quality RFA features improve the quality of image compositing of transparent objects on various backgrounds. As \cite{chen2018tom} introduced RFA for transparent object compositing, we propose to use this task as additional supervision. We introduce $L_{\text{comp}}$, as an additional supervisory loss, to further refine the estimated RFA intermediate features by the RFA feature regression network, thus improving the 6D pose estimation.
We use the estimated matte (i.e., RFA), for compositing the transparent object on a random background image for both groundtruth and predictions.
Drawing inspiration from \cite{chen2018tom} and \cite{zongker2023environment} for transparent object compositing, we apply the compositing equation to each pixel \cite{chen2018tom}:

\begin{equation}
    C = (1 - M_{\text{VIS}}) . B + M_{\text{VIS}} . M_{\text{RHO}} . f(\textbf{T}, M_{\text{FLOW}}) \label{eq:finalmatting}
\end{equation}

Here,  $C$ is the final computed compositing for a given pixel, $B$ is the background color, and $\textbf{T}$ are the set of calibration images \cite{chen2018tom}, and $f$ is the matting function. The variable $M_{\text{VIS}}$ in eq. \ref{eq:finalmatting} distinguishes between background and foreground.  Equation \ref{eq:finalmatting} is then estimated by using the groundtruth and estimated $M_{\text{VIS}}$, $M_{\text{RHO}}$, and $M_{\text{FLOW}}$ for each pixel of the transparent object. Fig. 3 shows examples of the final compositing of the transparent objects on random COCO backgrounds \cite{cocodataset}.

Thus we add $L_{\text{comp}}$ to the loss function~\ref{eq:FinalLoss}, redefining our final loss function $L$ as follows:

\begin{equation}
 L = L_{\text{inter}} + L_{\text{pose}} + L_{\text{comp}}  \label{eq:FinalLossWithMATT}
\end{equation}
Where
\begin{equation}
     L_{\text{comp}} = || \hat{M}_{\text{VIS}} \cdot \hat{C} -  \tilde{M}_{\text{VIS}} . \tilde{C} || 
\end{equation}

$\hat{C}$ is computed using groundtruth RFA and  $\tilde{C}$ is calculated using estimated RFA intermediate features during the training procedure using eq. \ref{eq:finalmatting}.

\section{Experiments}

In this section, we first introduce our experimental setup. We will detail the implementation of our method and the datasets being used, followed by the evaluation metrics. We then proceed to present the evaluation results for commonly employed benchmark transparent pose estimation methods against our method. 

\subsection{Experimental Setup}

\textbf{Implementation Details. }Our experiments are implemented using PyTorch \cite{paszke2019pytorch} and GDRNPP, the version of \cite{wang2021gdr} presented in the BOP challenge 2022 \cite{sundermeyer2023bop}. We train our networks on Nvidia-GTX 3090 and Nvidia A6000. Our method is trained end-to-end using the Ranger optimizer \cite{liu2019variance} \cite{zhang2019lookahead}, with a batch size of 8 and a base learning rate of 1e-4, which we anneal
at $72\%$ of the training phase using a cosine schedule \cite{loshchilov2016sgdr}. For transparent object compositing detailed in section 3. (b), we use COCO images \cite{cocodataset} as the random background.

\textbf{RFA matte rendering. }The training dataset is generated using BlenderProc\cite{Denninger2023}. Refractive flow, attenuation, and binary mask for each object under each viewpoint are obtained following the gray code-based calibration method used in~\cite{chen2018tom}. We keep the index-of-refraction fixed at $1.5$ for all objects.

\textbf{Datasets.}
We conduct our experiments on two datasets: \textit{TOD} (Transparent object dataset) which is introduced by the Keypose method~\cite{liu2020keypose}, and the Trans6D-32K introduced by TGF-Net~\cite{yu2023tgf}. The \textit{TOD} dataset provides 2D and 3D groundtruth keypoints and contains 15 unique transparent objects with varying shapes, scales, and symmetries. The dataset contains approximately 2700 training samples and 350 test samples per object, with each image sample of resolution 720$X$1080. For obtaining the groundtruth 6D pose from the provided groundtruth keypoints, we use the Orthogonal Procrustes algorithm \cite{gower1975generalized}, as done in \cite{liu2020keypose}. \textit{Trans6D-32K} contains 10 unique common types of household transparent objects, of which 5 are symmetric and 5 are non-symmetric objects. The dataset contains 400 synthetic training images and 2800 synthetic test images per object, so the entire dataset contains 32000 images.

\textbf{Evaluation Metrics.}
We use two standard metrics for 6D object pose evaluation, i.e. \textit{ADD(-S)} \cite{hodavn2016evaluation} \cite{hinterstoisser2013model} and \textit{Average Recall (AR)}\cite{hodan2018bop}. The ADD metric assesses whether the average deviation of transformed model points is within $10\%$ of the object's diameter (0.1d). For symmetric objects, ADD-S measures error as the average distance to the closest model point. Average Recall (AR) is computed as the mean of three metrics: Maximum Symmetry-Aware Projection Distance (MSPD), Maximum Symmetry-Aware Surface Distance (MSSD) and Visible Surface Discrepancy (VSD). For detailed explanations, please refer to \cite{hodavn2020bop}.

The Keypose method \cite{liu2020keypose} does not directly predict the 6D pose of the object, unlike our method, but instead estimates the 2D keypoints of the object. To ensure a fair comparison, we evaluate our method using Mean Absolute Error ($MAE$) scores for predicted keypoints. We leverage our trained 6D pose estimation model, and transform using the groundtruth keypoints provided by \textit{TOD}. Then we project it on the 2D image space of image $I$ using the camera intrinsics.

\subsection{Evaluation}

We evaluate our method against the Keypose \cite{liu2020keypose}, GDR-Net \cite{wang2021gdr} and TGF-Net \cite{yu2023tgf} method. For comparision with the Keypose method we train our method and GDR-Net with the $TOD$ $dataset$ and use the inference output per object, as provided by the trained Keypose models. For comparison with the TGF-Net, we train our method using the $Trans$$6D-32K$ dataset and use the evaluation results as provided in \cite{yu2023tgf}. The results of the comparison experiment trained with the $TOD$ $dataset$ are illustrated in Table I, and with the $Trans$$6D-32K$ dataset is illustrated in Table II.

\begin{figure*}[htp]
  \centering
  \includegraphics[width=0.92\textwidth]{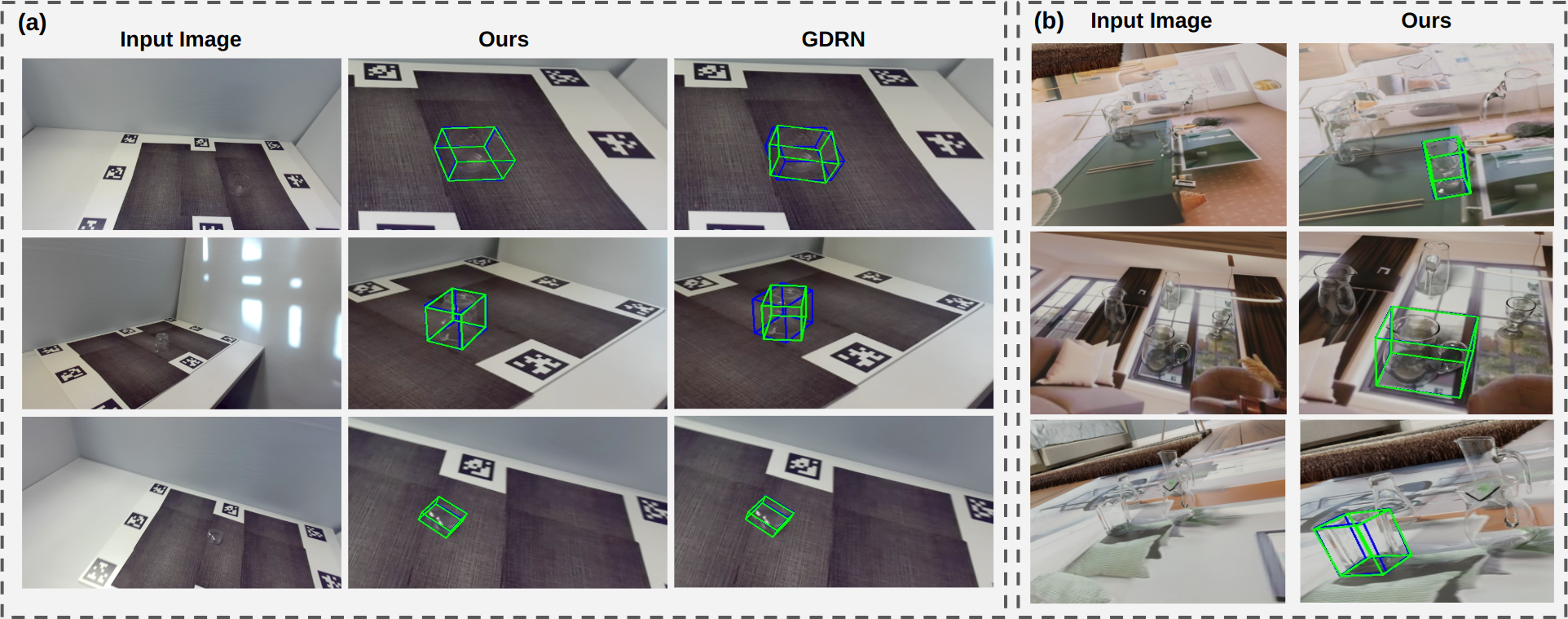}
  \caption{\textbf{Qualitative Results of ReFlow6D:} (a) Qualitative results on the TOD dataset. (b) Qualitative results on 
  the Trans32K-6D dataset. Estimates are shown in cropped images for visibility. No estimates are shown for the TGF-Net method as the authors did not publish their code.} 
\end{figure*}

\subsection{Comparison with the benchmark}

\begin{table}[ht!]
\centering
\caption{Average Recall results on the \textit{TOD} dataset objects}
\begin{tabular}{|c|c|c||c|c|c|}
\hline
Objects & \textbf{KeyPose} & \textbf{Ours} & \textbf{Keypose} & \textbf{GDR-Net} & \textbf{Ours} \\
\hline
\textbf{Metric} & \multicolumn{2}{|c||}{MAE$\downarrow$} & \multicolumn{3}{|c|}{AR$\uparrow$} \\
\hline
Bottle0 & \textbf{4.6} & 4.9 & \textbf{92.4} & 85.9 &  88.8\\
\hline
Bottle1 & 5.1 & \textbf{1.8} & 18.6 &  61.3 & \textbf{68.9} \\
\hline
Cup0 & 6.8 & \textbf{1.3} &  99.6 &  99.9 & \textbf{100.0} \\
\hline
Cup1 & 7.1 & \textbf{1.4} & 87.1 &  83.9 & \textbf{ 89.1} \\
\hline
Mug0 & \textbf{8.8} & 11.2 & 83.2 & 84.8 & \textbf{99.2} \\
\hline
Mug1 & 21.9 & \textbf{2.1} & 88.1 &  97.7 & \textbf{98.0} \\
\hline
Mug2 & 10.1 & \textbf{1.6} & 80.1 & \textbf{89.2} & 88.7 \\
\hline
Mug3 & 11.3 & \textbf{1.3} & 60.0 & 88.8 & \textbf{89.8} \\
\hline
Mug4 & 12.1 & \textbf{2.1} & 77.6 &  \textbf{91.7} & 88.8 \\
\hline
Mug5 & 9.0 & \textbf{1.9} & 89.2 &  \textbf{94.4} & 92.3 \\
\hline
Mug6 & 9.7 & \textbf{2.0} & 95.3 &  \textbf{98.6} & 98.2 \\
\hline
Tree0 & 15.6 & \textbf{6.1} & \textbf{91.7} &  88.8 & 89.3 \\
\hline
heart0 & 12.8 & \textbf{5.8} & 38.10 & 72.6 & \textbf{84.6} \\
\hline
\textbf{Mean} & 10.4 &  \textbf{3.4} & 77.0 &  87.5& \textbf{90.4} \\ 
\hline
\end{tabular}
\end{table}

\textbf{Results on TOD.}
Quantitative results are mentioned in Table 1. Our method achieves the best average recall of \textbf{90.4}\% against 87.5\% and 77.0\% for GDR-Net and Keypose method respectively. ReFlow6D outperforms the other two state-of-the-art methods for highly symmetric and textureless objects such as $Bottle1$, for which we achieve an AR score of \textbf{68.9}\%. For highly asymmetric objects such as $Tree0$, our method and GDR-Net method perform a bit worse than the Keypose method. This is because both the GDR-Net and our method rely on Surface Regions, which are difficult to predict for a complex geometry such as $Tree0$. Fig. 4(a) shows the qualitative results on TOD.\newline
For keypoint prediction, our method achieves the best \textit{MAE} score of \textbf{3.4} compared to Keypose's score of 10.4.\newline 
$Ball0$ was excluded due to the absence of a groundtruth binary mask, and $Bottle2$ was excluded due to inaccurate test-set groundtruth 6D poses provided by $TOD$. Thus only evaluating on 13 TOD dataset objects.

In addition to comparing MAE scores with the \cite{liu2020keypose} method, we also compare against MVTrans\cite{wang2023mvtrans}, a multi-view approach for transparent object perception and pose estimation. Our method achieves the best \textit{MAE} score of \textbf{3.4}, averaged over 13 TOD objects, compared to MVTrans's score of \textbf{7.4}.

\begin{table}
\centering
\caption{ADD(-S) on \textit{Trans6D-32K} dataset objects}
\begin{tabular}{|c|c|c|c|}
\hline
Objects & \textbf{GDR-Net} & \textbf{TGF-Net} & \textbf{Ours} \\
\hline
\#01 & 73.5 & 83.4 & \textbf{89.7} \\
\hline
\#02 & 76.3 & 83.5 & \textbf{95.2} \\
\hline
\#03 & 67.4 & 67.7 & \textbf{82.8} \\
\hline
\#04 & 82.9 & 85.4 & \textbf{91.0} \\
\hline
\#05 & 78.5 & 89.6 & \textbf{91.8} \\
\hline
\#06 & 91.5 & 89.6 & \textbf{94.7} \\
\hline
\#07 & 90.6 & 92.6 & \textbf{94.8} \\
\hline
\#08 & 96.6 & 97.3 & \textbf{99.3} \\
\hline
\#09 & 96.0 & 97.5 & \textbf{98.6} \\
\hline
\#10 & 92.4 & \textbf{94.9} & 93.8 \\
\hline
\hline
\textbf{Mean} & 84.6 & 88.2 & \textbf{93.2} \\
\hline
\end{tabular}
\end{table}

\textbf{Results on Trans6D-32K.}
We take the evaluation criteria i.e, the ADD(-s) and results report by the TGF-Net in their original paper \cite{yu2023tgf}. Our method outperforms TGF-Net and GDR-Net on the Trans6D-32K dataset, as indicated in Table 2. Notably, our method achieves the best ADD(-S) score of \textbf{$93.2$}, outperforming TGF-Net ($88.2$) and GDR-Net ($84.6$). Even complex non-symmetric objects like  $\#03$ \textit{ReFlow6D} perform much better with minimal degradation compared to the other methods.  Qualitative results on TOD are depicted in Fig. 4(b).

\subsection{Ablation Studies}
In order to verify the effectiveness of the RFA intermediate layers, we conducted ablation experiments. All ablation experiments use the same network initialization and training scheme. We consider the original GDR-Network \cite{wang2021gdr} without Dense correspondence maps as our base network since we only add the RFA intermediate layers to this.

\textbf{Effect without compositing Loss.}
We exclude the $L_{comp}$ detailed in eq. 9, while keeping the other losses and predicting all the RFA, $M_{SRA}$ and $M_{VIS}$. We can see from Table 3 that there is a decline in performance to $90.5$ from $93.2$ (as indicated in Table 2). Proving that compositing loss as an additional supervision loss contributes to the performance of our method. This also substantiates that improved RFA intermediate features contribute to improved accurate 6D pose estimation of transparent objects.

\textbf{Effect without Flow.}
We remove the prediction of Flow as intermediate layers. While keeping the prediction of other RFA, $M_{SRA}$, $M_{VIS}$ and losses. We can see in Table 3 the big performance drop to $80.3$ from $93.2$ in Table 1. This notable decrease in performance is primarily attributed to the crucial role played by refractive flow features in influencing the final 6D pose estimation, surpassing the contribution of other RFA features.

\textbf{Effect without Rho.}
We remove the prediction of attenuation $M_{RHO}$ as intermediate layers while keeping the prediction of other RFA components, $M_{SRA}$, $M_{VIS}$ and losses. From Table 3, we observe a marginal drop in performance to $93.0$ compared to the full model. This marginal drop is attributed to the fact that Flow along with Surface Regions encodes all the essential information required for predicting the 6D pose of the transparent object. This proves, that learning $M_{RHO}$ for pose estimation does not contribute greatly towards 6D pose estimation of the transparent objects, explaining the increase in performance when compared to the ablation studies of removing $L_{comp}$. Indicating that its absence facilitates easier learning for the Patch-PnP method.

We have opted not to present the ablation results for ReFlow6D trained exclusively with $M_{SRA}$. This is because the Patch-PnP network fails to converge when trained solely with $M_{SRA}$.

\begin{table}
\centering
\caption{Ablation study of RFA features in ADD(-S) on \textit{Trans6D-32K}}
\begin{tabular}{|c|c|c|c|c|c|}
\hline
\textbf{Features} & $L_{comp}$ & Flow & Rho & $M_{\text{SRA}}$ & ADD(-S) \\
\hline
w/o Compositing loss & $\times$ &$\checkmark$&$\checkmark$&$\checkmark$& 90.5 \\
\hline
w/o Flow & $\times$ & $\times$ &$\checkmark$&$\checkmark$& 80.3 \\
\hline
w/o Rho & $\times$ &$\checkmark$& $\times$ &$\checkmark$& 93.0 \\
\hline
\textbf{Full model} &$\checkmark$&$\checkmark$&$\checkmark$&$\checkmark$& \textbf{93.2} \\
\hline
\end{tabular}
\end{table}

\section{Real-world robot experiments}
\begin{figure*}[htp]
\centering
  \includegraphics[width=0.9\textwidth]
  {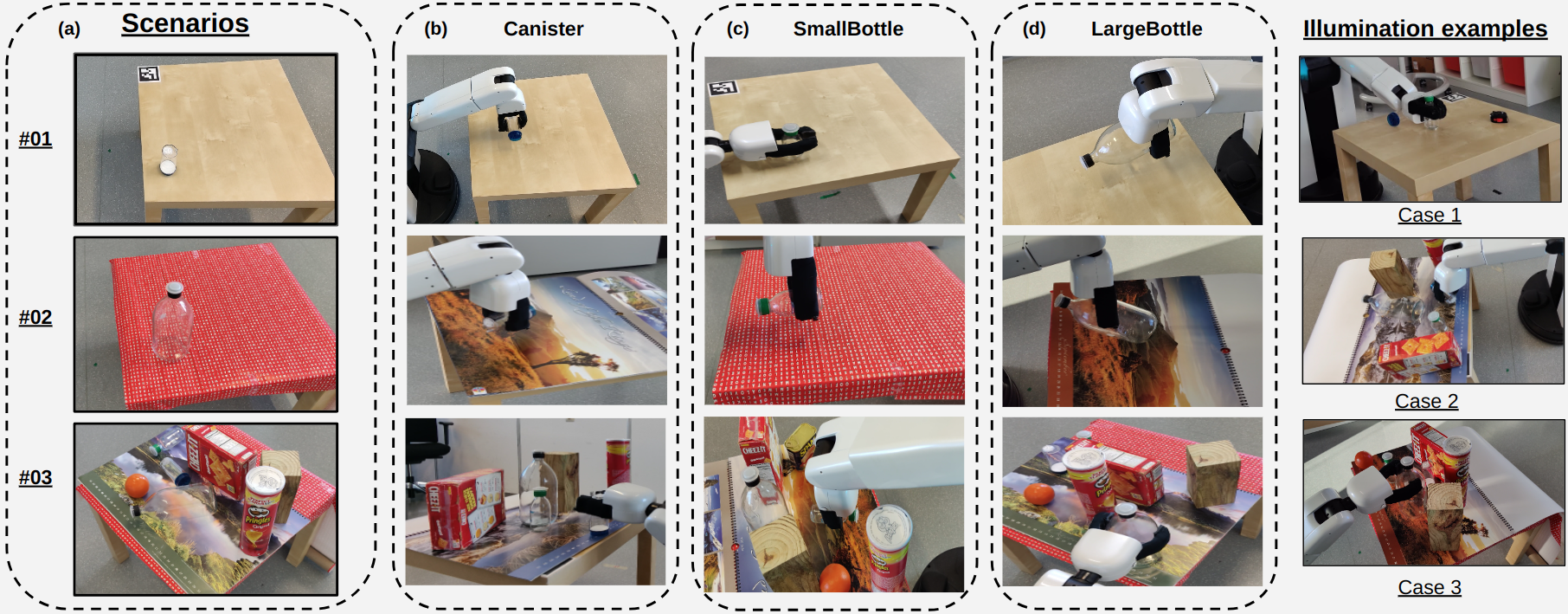}
  \caption{\textbf{Grasping qualitative results:} On the left we show (a) Examples of all three different scenarios. (b) Grasping example of the object "Canister" for all 3 scenarios. (c) Grasping example of the object "SmallBottle" for all 3 scenarios. (d) Grasping of the object "LargeBottle" for all 3 scenarios. On the right we show four cases of illumination (a) Light sifting through the semi-permeable blind covering the only window of the room. (b) Artificial ambient light. (c) Natural light. While the fourth case of illumination i.e, superposition of artificial ambient and natural light is shown on the left side of images with the grasping senarios. } 
  \label{fig:graspingQual}
\end{figure*}

To showcase that our proposed method ReFlow6D can be directly applied to real-world scenarios, we conduct robot grasping experiments.

\textbf{Dataset and Scene setup.} 
Due to the unavailability of the real physical object instances from the $TOD$ and $Trans32K-6D$ datasets for real robot grasping experiments, we use items from the TraceBot project~\footnote{H2020 program under grant agreement No 101017089}. 
We construct a physically based synthetic dataset of the available objects we term \emph{TraceBot dataset}.
The \emph{TraceBot dataset} includes a mixture of three transparent and six non-transparent objects as distractors for real-world grasping experiments. The three transparent objects are, the "$Canister$", "$SmallBottle$" and "$LargeBottle$". The object CAD models were constructed by precisely measuring each of the three objects by a caliper and constructing the model using SolidWorks.
To render this dataset we used Blender \cite{blender} with varied background and illumination for each image in the dataset, where all the nine objects were placed in random poses in the setup. 
For each of the three transparent objects, $5000$ images were rendered for training. \newline
The scene setup for the real-world grasping was set up in front of the robot on a table, as can be seen in Fig.~\ref{fig:graspingQual}. 
In order to create challenging and varying scene illumination the light is varied for each grasp. In particular we have consider four cases of illumination detailed in \ref{fig:graspingQual}.

In total, we created and test three scenarios with considerable domain shift to the training data. The three scenarios can be described as follows:

\begin{itemize}
    \item \textbf{Scenario 1} includes only the raw table plane unseen in the training data, with the single transparent object placed randomly on it. An qualitative example for scenario 1 can be seen in Fig.~\ref{fig:graspingQual}(a)($01$).
    \item \textbf{Scenario 2} includes a random textured background unseen in the training data, placed on the table for each grasp. The single transparent object is also placed randomly on it for each real-robot grasp experiment. The qualitative example for scenario 2 can be seen in Fig.~\ref{fig:graspingQual}(a)($02$).
    \item \textbf{Scenario 3} is a cluttered scene consisting of the two other transparent objects with 5 other randomly chosen unseen opaque objects, together with unseen random background texture for each grasp. Both the background texture and the opaque objects were not part of the training data. All the objects are randomly placed on the table with random backgrounds. The qualitative example for scenario 3 can be seen in Fig.~\ref{fig:graspingQual}(a)($03$).
\end{itemize}

\begin{table}
\centering
\caption{Grasping experiments results}
\begin{tabular}{|c|c|c|c|c|}
\hline
                       & Canister & SmallBottle & LargeBottle & \textbf{Mean} \\
\hline
Scenario 01          & 90\% & 100\% & 80\% & 90\% \\
\hline
Scenario 02          & 90\% & 90\% & 80\% & 86.6\% \\
\hline
Scenario 03          & 80\% & 70\% & 60\% & 70\%\\
\hline
\textbf{Total Success} & 86.6\% & 86.6\% & 73.3\% & 82.2\%\\
\hline
\end{tabular}
\end{table}

\textbf{Hardware and Implementation.}
We employ the Toyota HSR robot \cite{yamamoto2018development} which comes with the RGB-D camera Xtion PRO LIVE mounted, for all our grasping experiments.
We train ReFlow6D network following the implementation details mentioned in Section 4. Fixed grasp points were manually selected and annotated based on the CAD model of the objects and the robot's gripper in advance. We first train a YOLOv3 \cite{redmon2018yolov3} detection network to
detect the 2D bounding box of the object, after which we trained the ReFlow6D method for 6D pose estimation per object. Finally, we used the robot to grasp transparent objects to demonstrate the applicability of our method in real-world scenarios. We used the same grasp annotation method and grasping pipeline as described in \cite{gupta2022grasping}. We conducted a total of 30 grasps per object, i.e., 10 grasps for each of the mentioned scenarios.

\textbf{Results.}
In the qualitative experimental results fig.\ref{fig:graspingQual}(b)(c)(d), we showcase performance for each object and scenario. Quantitative results in Table 4 detail 10 grasps per object per scenario, yielding a mean success rate of \textbf{82.2\%}. $LargeBottle$ exhibits the lowest success rate \textbf{73.3\%} success rate for 30 grasps, attributed to its size similarity with our robot's gripper. This highlights the need for precise grasp-planning, object detection, and 6D pose estimation, leaving a smaller margin of error. These results also highlight how ReFlow6D's accurate pose estimation translates effectively to real-world tasks.

\section{Conclusion}

In this work, we propose ReFlow6D, a monocular instance-level 6D pose estimation approach tailored specifically for transparent objects. Our method proposes a novel set of \textit{refractive-intermediate representations}, guided by refractive principles and enabling robust transparent object pose estimation. We demonstrated that integrating these refractive feature attributes (RFAs) alongside surface-region attention as intermediate features better guide the network toward more precise 6D pose estimations for transparent objects. Through comprehensive empirical evaluations, we demonstrate the effectiveness of ReFlow6D in real-world scenarios compared to existing state-of-the-art methods. These results underscore the efficacy of the \textit{refractive-intermediate representation} over geometric and edge-based representations. Future work will investigate pose estimation of objects with more complex transparent objects with diverse thickness, geometry, and indices of refraction.

\bibliographystyle{IEEEtran}
\bibliography{blibli}

\begin{thebibliography}{10}
\providecommand{\url}[1]{#1}
\csname url@samestyle\endcsname
\providecommand{\newblock}{\relax}
\providecommand{\bibinfo}[2]{#2}
\providecommand{\BIBentrySTDinterwordspacing}{\spaceskip=0pt\relax}
\providecommand{\BIBentryALTinterwordstretchfactor}{4}
\providecommand{\BIBentryALTinterwordspacing}{\spaceskip=\fontdimen2\font plus
\BIBentryALTinterwordstretchfactor\fontdimen3\font minus \fontdimen4\font\relax}
\providecommand{\BIBforeignlanguage}[2]{{%
\expandafter\ifx\csname l@#1\endcsname\relax
\typeout{** WARNING: IEEEtran.bst: No hyphenation pattern has been}%
\typeout{** loaded for the language `#1'. Using the pattern for}%
\typeout{** the default language instead.}%
\else
\language=\csname l@#1\endcsname
\fi
#2}}
\providecommand{\BIBdecl}{\relax}
\BIBdecl

\bibitem{thalhammer2021pyrapose}
S.~Thalhammer, M.~Leitner, T.~Patten, and M.~Vincze, ``Pyrapose: Feature pyramids for fast and accurate object pose estimation under domain shift,'' in \emph{2021 IEEE International Conference on Robotics and Automation (ICRA)}.\hskip 1em plus 0.5em minus 0.4em\relax IEEE, pp. 13\,909--13\,915.

\bibitem{stevvsic2020learning}
S.~Stev{\v{s}}i{\'c}, S.~Christen, and O.~Hilliges, ``Learning to assemble: Estimating 6d poses for robotic object-object manipulation,'' \emph{IEEE Robotics and Automation Letters}, vol.~5, no.~2, pp. 1159--1166, 2020.

\bibitem{hodavn2016evaluation}
T.~Hoda{\v{n}}, J.~Matas, and {\v{S}}.~Obdr{\v{z}}{\'a}lek, ``On evaluation of 6d object pose estimation,'' in \emph{Computer Vision--ECCV 2016 Workshops: Amsterdam, The Netherlands, October 8-10 and 15-16, 2016, Proceedings, Part III 14}.\hskip 1em plus 0.5em minus 0.4em\relax Springer, 2016, pp. 606--619.

\bibitem{hodan2017t}
T.~Hodan, P.~Haluza, {\v{S}}.~Obdr{\v{z}}{\'a}lek, and et.al, ``T-less: An rgb-d dataset for 6d pose estimation of texture-less objects,'' in \emph{2017 IEEE Winter Conference on Applications of Computer Vision (WACV)}.\hskip 1em plus 0.5em minus 0.4em\relax IEEE, 2017, pp. 880--888.

\bibitem{wu2021pseudo}
C.~Wu, L.~Chen, Z.~He, and J.~Jiang, ``Pseudo-siamese graph matching network for textureless objects’6-d pose estimation,'' \emph{IEEE Transactions on Industrial Electronics}, vol.~69, no.~3, pp. 2718--2727, 2021.

\bibitem{hodan2020epos}
T.~Hodan, D.~Barath, and J.~Matas, ``Epos: Estimating 6d pose of objects with symmetries,'' in \emph{Proceedings of the IEEE/CVF conference on computer vision and pattern recognition}, 2020, pp. 11\,703--11\,712.

\bibitem{richter2021handling}
J.~Richter-Klug and U.~Frese, ``Handling object symmetries in cnn-based pose estimation,'' in \emph{IEEE International Conference on Robotics and Automation (ICRA)}.\hskip 1em plus 0.5em minus 0.4em\relax IEEE, 2021, pp. 13\,850--13\,856.

\bibitem{peng2019pvnet}
S.~Peng, Y.~Liu, and et.al, ``Pvnet: Pixel-wise voting network for 6dof pose estimation,'' in \emph{Proceedings of the IEEE/CVF Conference on Computer Vision and Pattern Recognition}, 2019, pp. 4561--4570.

\bibitem{chang2021ghostpose}
J.~Chang, M.~Kim, S.~Kang, and et.al, ``Ghostpose: Multi-view pose estimation of transparent objects for robot hand grasping,'' in \emph{IEEE/RSJ International Conference on Intelligent Robots and Systems (IROS)}.\hskip 1em plus 0.5em minus 0.4em\relax IEEE, 2021, pp. 5749--5755.

\bibitem{lysenkov2013pose}
I.~Lysenkov and V.~Rabaud, ``Pose estimation of rigid transparent objects in transparent clutter,'' in \emph{IEEE International Conference on Robotics and Automation}.\hskip 1em plus 0.5em minus 0.4em\relax IEEE, 2013, pp. 162--169.

\bibitem{sajjan2020clear}
S.~Sajjan, M.~Moore, and e.~a. Pan, ``Clear grasp: 3d shape estimation of transparent objects for manipulation,'' in \emph{IEEE International Conference on Robotics and Automation (ICRA)}.\hskip 1em plus 0.5em minus 0.4em\relax IEEE, 2020, pp. 3634--3642.

\bibitem{chen2022clearpose}
X.~Chen, H.~Zhang, Z.~Yu, A.~Opipari, and O.~Chadwicke~Jenkins, ``Clearpose: Large-scale transparent object dataset and benchmark,'' in \emph{European Conference on Computer Vision}.\hskip 1em plus 0.5em minus 0.4em\relax Springer, 2022, pp. 381--396.

\bibitem{zhu2021rgb}
L.~Zhu, A.~Mousavian, Y.~Xiang, and et.al, ``Rgb-d local implicit function for depth completion of transparent objects,'' in \emph{Proceedings of the IEEE/CVF Conference on Computer Vision and Pattern Recognition}, 2021, pp. 4649--4658.

\bibitem{liu2020keypose}
X.~Liu, R.~Jonschkowski, A.~Angelova, and K.~Konolige, ``Keypose: Multi-view 3d labeling and keypoint estimation for transparent objects,'' in \emph{Proceedings of the IEEE/CVF conference on computer vision and pattern recognition}, 2020, pp. 11\,602--11\,610.

\bibitem{wang2021gdr}
G.~Wang, F.~Manhardt, F.~Tombari, and X.~Ji, ``Gdr-net: Geometry-guided direct regression network for monocular 6d object pose estimation,'' in \emph{Proceedings of the IEEE/CVF Conference on Computer Vision and Pattern Recognition}, 2021, pp. 16\,611--16\,621.

\bibitem{hodan2018bop}
T.~Hodan, F.~Michel, E.~Brachmann, and et.al, ``Bop: Benchmark for 6d object pose estimation,'' in \emph{Proceedings of the European conference on computer vision (ECCV)}, 2018, pp. 19--34.

\bibitem{yu2023tgf}
H.~Yu, S.~Li, H.~Liu, and et.al, ``Tgf-net: Sim2real transparent object 6d pose estimation based on geometric fusion,'' \emph{IEEE Robotics and Automation Letters}, 2023.

\bibitem{chen2018tom}
G.~Chen, K.~Han, and K.-Y.~K. Wong, ``Tom-net: Learning transparent object matting from a single image,'' in \emph{Proceedings of the IEEE conference on computer vision and pattern recognition}, 2018, pp. 9233--9241.

\bibitem{zongker2023environment}
D.~E. Zongker, D.~M. Werner, B.~Curless, and D.~H. Salesin, ``Environment matting and compositing,'' in \emph{Seminal Graphics Papers: Pushing the Boundaries, Volume 2}, 2023, pp. 537--546.

\bibitem{lai2015transparent}
P.-J. Lai and C.-S. Fuh, ``Transparent object detection using regions with convolutional neural network,'' in \emph{IPPR conference on computer vision, graphics, and image processing}, vol.~2, 2015.

\bibitem{xie2020segmenting}
E.~Xie, W.~Wang, W.~Wang, M.~Ding, C.~Shen, and P.~Luo, ``Segmenting transparent objects in the wild,'' in \emph{Computer Vision--ECCV 2020: 16th European Conference, Glasgow, UK, August 23--28, 2020, Proceedings, Part XIII 16}.\hskip 1em plus 0.5em minus 0.4em\relax Springer, 2020, pp. 696--711.

\bibitem{kalra2020deep}
A.~Kalra, V.~Taamazyan, S.~K. Rao, and et.al, ``Deep polarization cues for transparent object segmentation,'' in \emph{Proceedings of the IEEE/CVF Conference on Computer Vision and Pattern Recognition}, 2020, pp. 8602--8611.

\bibitem{ichnowski2021dex}
J.~Ichnowski, Y.~Avigal, J.~Kerr, and K.~Goldberg, ``Dex-nerf: Using a neural radiance field to grasp transparent objects,'' \emph{arXiv preprint arXiv:2110.14217}, 2021.

\bibitem{wang2019densefusion}
C.~Wang, D.~Xu, Y.~Zhu, and et.al, ``Densefusion: 6d object pose estimation by iterative dense fusion,'' in \emph{Proceedings of the IEEE/CVF conference on computer vision and pattern recognition}, 2019, pp. 3343--3352.

\bibitem{zhang2022transnet}
H.~Zhang, A.~Opipari, and et.al, ``Transnet: Category-level transparent object pose estimation,'' in \emph{European Conference on Computer Vision}.\hskip 1em plus 0.5em minus 0.4em\relax Springer, 2022, pp. 148--164.

\bibitem{li2019cdpn}
Z.~Li, G.~Wang, and X.~Ji, ``Cdpn: Coordinates-based disentangled pose network for real-time rgb-based 6-dof object pose estimation,'' in \emph{Proceedings of the IEEE/CVF International Conference on Computer Vision}, 2019, pp. 7678--7687.

\bibitem{chuang2000environment}
Y.-Y. Chuang, D.~E. Zongker, and et.al, ``Environment matting extensions: Towards higher accuracy and real-time capture,'' in \emph{Proceedings of the 27th annual conference on Computer graphics and interactive techniques}, 2000, pp. 121--130.

\bibitem{tian2019fcos}
Z.~Tian, C.~Shen, H.~Chen, and T.~He, ``Fcos: Fully convolutional one-stage object detection,'' in \emph{Proceedings of the IEEE/CVF international conference on computer vision}, 2019, pp. 9627--9636.

\bibitem{redmon2018yolov3}
J.~Redmon and A.~Farhadi, ``Yolov3: An incremental improvement,'' \emph{arXiv preprint arXiv:1804.02767}, 2018.

\bibitem{su2022zebrapose}
Y.~Su, M.~Saleh, T.~Fetzer, and et.al, ``Zebrapose: Coarse to fine surface encoding for 6dof object pose estimation,'' in \emph{Proceedings of the IEEE/CVF Conference on Computer Vision and Pattern Recognition}, 2022, pp. 6738--6748.

\bibitem{cocodataset}
\BIBentryALTinterwordspacing
T.~Lin, M.~Maire, S.~J. Belongie, and et.al, ``Microsoft {COCO:} common objects in context,'' \emph{CoRR}, vol. abs/1405.0312, 2014. [Online]. Available: \url{http://arxiv.org/abs/1405.0312}
\BIBentrySTDinterwordspacing

\bibitem{paszke2019pytorch}
A.~Paszke, S.~Gross, F.~Massa, and et.al, ``Pytorch: An imperative style, high-performance deep learning library,'' \emph{Advances in neural information processing systems}, vol.~32, 2019.

\bibitem{sundermeyer2023bop}
M.~Sundermeyer, T.~Hoda{\v{n}}, and et.al, ``Bop challenge 2022 on detection, segmentation and pose estimation of specific rigid objects,'' in \emph{Proceedings of the IEEE/CVF Conference on Computer Vision and Pattern Recognition}, 2023, pp. 2785--2794.

\bibitem{liu2019variance}
L.~Liu, H.~Jiang, and et.al, ``On the variance of the adaptive learning rate and beyond,'' \emph{arXiv preprint arXiv:1908.03265}, 2019.

\bibitem{zhang2019lookahead}
M.~Zhang, J.~Lucas, J.~Ba, and G.~E. Hinton, ``Lookahead optimizer: k steps forward, 1 step back,'' \emph{Advances in neural information processing systems}, vol.~32, 2019.

\bibitem{loshchilov2016sgdr}
I.~Loshchilov and F.~Hutter, ``Sgdr: Stochastic gradient descent with warm restarts,'' \emph{arXiv preprint arXiv:1608.03983}, 2016.

\bibitem{Denninger2023}
\BIBentryALTinterwordspacing
M.~Denninger, D.~Winkelbauer, and et.al, ``Blenderproc2: A procedural pipeline for photorealistic rendering,'' \emph{Journal of Open Source Software}, vol.~8, no.~82, p. 4901, 2023. [Online]. Available: \url{https://doi.org/10.21105/joss.04901}
\BIBentrySTDinterwordspacing

\bibitem{gower1975generalized}
J.~C. Gower, ``Generalized procrustes analysis,'' \emph{Psychometrika}, vol.~40, pp. 33--51, 1975.

\bibitem{hinterstoisser2013model}
S.~Hinterstoisser, V.~Lepetit, and et.al, ``Model based training, detection and pose estimation of texture-less 3d objects in heavily cluttered scenes,'' in \emph{Computer Vision--ACCV 2012: 11th Asian Conference on Computer Vision, Daejeon, Korea, November 5-9, 2012, Revised Selected Papers, Part I 11}.\hskip 1em plus 0.5em minus 0.4em\relax Springer, 2013, pp. 548--562.

\bibitem{hodavn2020bop}
T.~Hoda{\v{n}}, M.~Sundermeyer, B.~Drost, and et.al, ``Bop challenge 2020 on 6d object localization,'' in \emph{Computer Vision--ECCV 2020 Workshops: Glasgow, UK, August 23--28, 2020, Proceedings, Part II 16}.\hskip 1em plus 0.5em minus 0.4em\relax Springer, 2020, pp. 577--594.

\bibitem{wang2023mvtrans}
Y.~R. Wang, Y.~Zhao, H.~Xu, and et.al, ``Mvtrans: Multi-view perception of transparent objects,'' in \emph{2023 IEEE International Conference on Robotics and Automation (ICRA)}.\hskip 1em plus 0.5em minus 0.4em\relax IEEE, 2023, pp. 3771--3778.

\bibitem{blender}
\BIBentryALTinterwordspacing
B.~O. Community, \emph{Blender - a 3D modelling and rendering package}, Blender Foundation, Stichting Blender Foundation, Amsterdam, 2018. [Online]. Available: \url{http://www.blender.org}
\BIBentrySTDinterwordspacing

\bibitem{yamamoto2018development}
T.~Yamamoto, K.~Terada, A.~Ochiai, and et.al, ``Development of the research platform of a domestic mobile manipulator utilized for international competition and field test,'' in \emph{IEEE/RSJ International Conference on Intelligent Robots and Systems (IROS)}.\hskip 1em plus 0.5em minus 0.4em\relax IEEE, 2018, pp. 7675--7682.

\bibitem{gupta2022grasping}
H.~Gupta, S.~Thalhammer, M.~Leitner, and M.~Vincze, ``Grasping the inconspicuous,'' \emph{arXiv preprint arXiv:2211.08182}, 2022.

\end{thebibliography}

\end{document}